# A Review of the Marathi Natural Language Processing

**Asang Dani** and **Dr. Shailesh R Sathe**
Department of Computer Science & Engineering
Visvesvaraya National Institute of Technology Nagpur, India
dt23cse001@students.vnit.ac.in and srsathe@cse.vnit.ac.in

## Abstract

Marathi is one of the most widely used languages in the world. One might expect that the latest advances in NLP research in languages like English reach such a large community. However, NLP advancements in English didn't immediately reach Indian languages like Marathi. There were several reasons for this. They included diversity of scripts used, lack of (publicly available) resources like tokenization strategies, high quality datasets & benchmarks, and evaluation metrics. In addition to this, the morphologically rich nature of Marathi, made NLP tasks challenging. Advances in Neural Network (NN) based models and tools since the early 2000s helped improve this situation and make NLP research more accessible. In the past 10 years, significant efforts were made to improve language resources for all 22 scheduled languages of India. This paper presents a broad overview of evolution of NLP research in Indic languages with a focus on Marathi and state-of-the-art resources and tools available to the research community. It also provides an overview of tools & techniques associated with Marathi NLP tasks.

## 1 Introduction

Natural Language Processing (NLP) has been an active area of research since early 1970s. Initial research included Machine Translation (MT), parsing, syntax analysis (Earley, 1970) and Morphological analysis (Koskenniemi, 1983). With the advent of faster computers, early rule based systems were slowly replaced by Statistical Machine Translation (SMT) (Brown et al., 1990). This allowed some early work in Speech Recognition and Information Retrieval (IR) (Salton and Buckley, 1988). During 1990s, in addition to SMT, there was ongoing work in other areas like Part-of-Speech (POS) Tagging (Brill, 1992), Named Entity Recognition (NER) (Grishman and Sundheim, 1996) and Natural Language Generation (NLG) (Reiter and Dale, 1997).

Tokenization is the first step in all NLP tasks. The text is broken into smaller units. It has evolved from simple rule-based methods to sophisticated subword tokenization techniques like Byte-Pair Encoding (BPE) (Sennrich et al., 2016), WordPiece (Wu et al., 2016), and SentencePiece (Kudo and Richardson, 2018), which are crucial for modern transformer-based (Vaswani et al., 2017) language models. Each advancement in tokenization has contributed to the efficiency, robustness, and the effectiveness of NLP models for various languages and tasks.

Word representations like (Mikolov et al., 2013b) demonstrated that continuous space representations, learned by neural networks, exhibit linguistic regularities, such as the ability to capture analogical relationships between words. Word2Vec (Mikolov et al., 2013a) provided a formal approach for building rich word embedding models for language learning tasks. Other tokenization approaches include GloVe (Pennington et al., 2014), ELMo (Peters et al., 2018) etc. They enable understanding the contextualized meaning of words/tokens.

Early 2000s saw resurgence of research in Artificial Neural Networks (ANNs) due to multiple factors like the availability of large datasets for training, faster CPUs and GPUs. This period saw significant advances in NLP research. Recurrent Neural Networks (RNNs) based models like Long Short Term Memory (LSTM) (Hochreiter and Schmidhuber, 1997) and Gated Recurrent Unit (GRU)(Cho et al., 2014a), allowed tracking long range dependencies. Neural Machine Translation (NMT) research made rapid strides with Sequence-to-Sequence models (Sutskever et al., 2014) and encoder-decoder architectures (Cho et al., 2014b), (Chung et al., 2014). Attention mechanism introduced by (Bahdanau et al., 2015) further improved Sequence-to-Sequence models. Self-Attention mechanism proposed by (Vaswani et al., 2017) is one of the most influential works that became the foundation of most modern LLMs in use today. This is popularly known as the Transformer model. Bidirectional Encoder Representations from Transformers (BERT) (Devlin et al., 2018) is one of the most widely used Transformers based model today for various language tasks. It is especially good at Natural Language Understanding (NLU). Bidirectional and Auto-Regressive Transformer (BART) (Lewis et al., 2020) which is a denoising autoencoder for pre-training sequence-to-sequence models is popular for Natural Language Generation (NLG) tasks.

India is very rich in its linguistic diversity (Emeneau, 1956). Early NLP research Indic languages focused on syntax and morphological analysis (Suryawanshi et al., 1994). Subsequent research focused on building lexical resources and corpora for Indic languages (Bhattacharyya, 2004). SMT based systems for Marathi were also developed in the late 2000s (Bhattacharyya and Panchanathan, 2008). It took some time for the advances in LLMs based on transformer architecture to reach Indic languages. One of the biggest obstacles was the lack of quality resources and datasets available to researchers. More work was also required on development of benchmarks for evaluating the results of Indic language tasks as metrics for English don't always work well for South Asian Languages. This paper covers the evolution of Indic NLP research and tools with a focus on Marathi.

## 2 Related Work

A Survey on NLP Resources, Tools, and Techniques for Marathi Language Processing (Lahoti et al., 2022), provides a good introduction to the Marathi language and its characteristics. It provides an overview of various monolingual and parallel corpora available for Marathi NLP. In addition to the text, speech corpora have also been included in the study. This overview of NLP processing techniques covers the evolution from Rule-based techniques to Neural Network based ones. However, it does not cover modern DL based approaches and different metrics used for evaluating the performance of Marathi NLP tasks.

## 3 Processing pipeline for NN based NLP systems

Figure 1 shows the pipeline for typical NLP systems based on neural networks. Step 7 is task specific and will be different for NLP tasks like NMT, NER, Abstractive Summarization, etc.

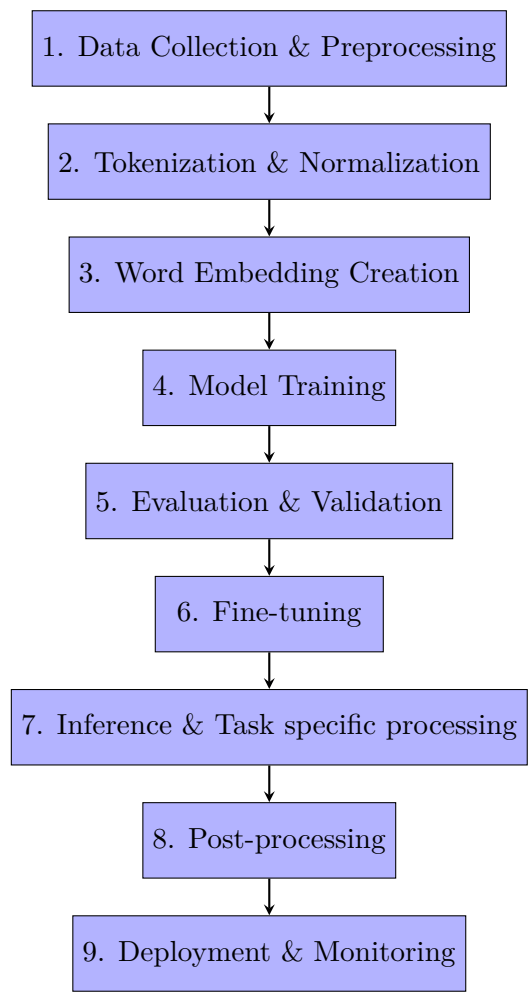

Figure 1: Pipeline for DL based NLP systems

## 4 Training Data and Benchmarks

There are two types of Corpora that are widely used for NLP tasks:

**Monolingual Corpora** A monolingual corpus consists of texts in a single language.

It helps a Deep Learning (DL) language model to understand vocabulary, sentence structure, phrases and morphology, text generation etc.

**Parallel Corpora** A parallel corpus consists of texts in one language aligned with their translations in another language (or several languages). This is extremely useful for tasks like Machine Translation, Lexicon building, etc.

Indic languages use a wide variety of scripts. The Central Institute of Indian Languages (CIIL)[1] started collecting textual data for various Indian languages in the 1970s. These early efforts were crucial in documenting and digitizing Indian languages, though the corpora were limited in scope and size. Text Encoding Initiative (TEI) [2] started developing a standard for the representation of texts in digital form. Early Indic corpora efforts adopted it. EMILLE Project (Enabling Minority Language Engineering) (Baker et al., 2004) was undertaken by Universities of Lancaster and Sheffield in collaboration with CIIL. The primary resource developed by the project is the EMILLE Corpus. It contains monolingual corpora for 14 South Asian languages totaling more than 96 million words. It also contains a parallel corpus of English and five of these languages. It also has part-of-speech (POS) tagged Urdu data. It contains 2.2 million words in Marathi monolingual corpus.

Another notable resource is Indo WordNet (Bhattacharyya, 2010). WordNets are lexical structures composed of synsets (sets of synonyms) and semantic relations. Semantic relations like hypernymy (is-a), meronymy (part-of), and troponymy (manner-of) link them. It is based on English WordNet built at the Princeton University (Miller, 1995). Starting with Hindi, it was expanded to include 18 Indian languages. Currently[3], it contains a total of 32,830 Marathi words.

The Indian Language Corpora Initiative (ILCI) (Jha, 2010) project, funded by the Department of Information Technology, Government of India, focuses on creating parallel corpora for major Indian languages. The primary objective was to develop linguistic resources to aid machine translation and other NLP tasks. This corpora was recently expanded (Siripragada et al., 2020) to include 407K aligned sentences across 10 Indic languages and relies on two sources the Press Information Bureau (PIB) and Mann Ki Baat[4], the Indian Prime Ministers speeches.

The Open Super-large Crawled ALMAnaCH coRpus (OSCAR) corpora (Suárez et al., 2019) are large-scale multilingual text corpora extracted from Common Crawl web data[5]. They are designed to support natural language processing (NLP) and linguistic research for a wide range of languages. The OSCAR project aims to provide freely accessible, high-quality text data for many under-resourced languages. It covers 166 languages and has a total of 82 million Marathi words.

IndicNLP Corpus (Kunchukuttan et al., 2020) is a large-scale web-based corpus for Indian languages developed using web crawling and data mining techniques. It contains 142 million Marathi words and a total of 2.7 billion words for 10 Indian languages from two language families and is designed for modern NLP applications like language modeling, sentiment analysis, and more. It draws on OSCAR Corpus as one of the sources and improves it significantly.

Samanantar (Ramesh et al., 2022) is a large parallel corpora for Indic languages. It includes 49.7 million sentence pairs between English and 11 Indic languages (from two language families). In addition, it has 83.4 million sentence pairs between all 55 Indic language pairs from the English-centric parallel corpus using English as the pivot language. The sentences are chosen from diverse domains. To ascertain the quality of the sentences from the corpus, 38 qualified annotators produced a total of 9,566 sentence pairs across 11 languages. This proves that the quality of corpora was a major focus area for the authors.

FLORES-200 (Few-shot Language Organization Resources Evaluation Setup) was introduced as a part of NLLB (No Language Left Behind) (Costa-jussà et al., 2022). It sup-ports 200 languages and is an extension to earlier FLORES-101 (Goyal et al., 2021). The dataset is curated with high-quality translations. It consists of 3001 sentences sampled from English-language Wikimedia projects for 204 total languages. NLLB-Seed consists of around six thousand sentences in 39 languages.

---

[1] https://ciil.org/
[2] https://tei-c.org/
[3] https://www.cfilt.iitb.ac.in/indowordnet/home#currentStatistics
[4] https://www.narendramodi.in/mann-ki-baat/
[5] https://commoncrawl.org/

OPUS Collection (Tiedemann, 2012) is one of the most popular resource for parallel corpora. As per the latest list[6], NLLB (Costa-jussà et al., 2022) is the largest aligned corpora for English to Marathi and Marathi to English in terms of number of tokens and sentences.

# 5 Tokenization

In almost all ML and DL algorithms for NLP, the first step is tokenization. It is the process of breaking down the text into smaller units, which could be words, subwords, characters, or even larger units such as phrases. It enables the analysis of text data by converting unstructured text into a structured form that the machine learning models can process. Early rule based tokenization was straightforward for English and other languages with clear word boundaries, typically marked by spaces. Early tokenizers simply split text at whitespace and punctuation, assuming that words are meaningful units. For languages with a rich morphological structure like Marathi, tokenization requires more sophisticated approaches to handle compound words and inflections. Subsequently, many rule based models that rely on linguistic understanding evolved. Statistical methods, such as Hidden Markov Models, were introduced to learn the likelihood of token boundaries from annotated corpora. These models could better handle ambiguous cases, such as distinguishing between "New York" as a single entity versus "New" and "York" as separate words.

Subword tokenization methods became more commonplace with the rise of NN-based models, especially for handling large vocabularies and rare words. Techniques like Byte-Pair Encoding (BPE) (Sennrich et al., 2016) and WordPiece (Wu et al., 2016) involve breaking down words into smaller units, which allows for the representation of out-of-vocabulary (OOV) words and reduces the size of the vocabulary that models need to handle.

BPE uses ideas from a compression algorithm (Gage, 1994) for segmenting the words. It initializes the symbol vocabulary with the character vocabulary, and represents each word as a sequence of characters, plus a special end-of-word symbol '·', which allows us to restore the original tokenization after translation. It iteratively counts all symbol pairs and replaces each occurrence of the most frequent pair ('A', 'B') with a new symbol 'AB'. Each merge operation produces a new symbol which represents a character n-gram. This process continues until we reach a desired vocabulary size. This technique enables Open Vocabulary translation which is vital for morphologically rich languages like Marathi.

WordPiece (Wu et al., 2016) is a popular subword segmentation algorithm. It breaks the words into subwords, which is then used to build a vocabulary. The initial vocabulary starts with individual characters of the language and common subwords are added based on their frequency in the training corpus. Starting with a training corpus and a number of desired tokens, it chooses the optimal wordpieces to ensure that the resulting corpus has minimal number of wordpieces when segmented according to the chosen wordpiece model (Luong et al., 2015).

SentencePiece (Kudo and Richardson, 2018) is a language independent subword tokenizer and detokenizer designed for text processing tasks. It doesn't require inputs to be pre-tokenized and can be directly trained from raw sentences. It not only performs subword tokenization, but converts the text into id sequence, which helps to develop a purely end-to-end system without relying on language specific resources. The main emphasis is on moving towards language-agnostic architecture. It also supports on-the-fly processing as is several orders of magnitude faster than most of the earlier systems. It is used with T5 (Text-to-Text Transfer Transformer)[7] based models (Raffel et al., 2019).

---

[6]https://opus.nlpl.eu/results/en&mr/corpus-result-table

[7]https://github.com/google-research/text-to-text-transfer-transformer

# 6 Evolution of NN Models for NLP tasks

Recurrent Neural Networks (RNNs) based models like Long Short Term Memory (LSTM) (Hochreiter and Schmidhuber, 1997) and Gated Recurrent Unit (GRU)(Cho et al., 2014a) allowed tracking long range dependencies. Neural Machine Translation (NMT) research made rapid strides with Sequence to sequence models (Sutskever et al., 2014) and encoder-decoder architectures (Cho et al., 2014b), (Chung et al., 2014). Attention mechanism (Bahdanau et al., 2015) further improved Sequence-to-Sequence models. Self-Attention mechanism (Vaswani et al., 2017) is one of the most influential works that became the foundation of most modern LLMs in use today. It is also called as the Transformer model. BERT as well as BART based architectures have become quite popular with NLP community and have been used for NLP tasks.

## 6.1 BERT

Bidirectional Encoder Representations from Transformers (BERT) (Devlin et al., 2018) and its variants are some of the most widely used language model today for various tasks. It is also known as mBERT or multilingual-BERT. BERT is designed to pre-train deep bidirectional representations from unlabeled text by jointly conditioning on both left and right context in all layers. A pre-trained model is typically fine-tuned with an additional output layer to create task specific models. A BERT model only uses the Encoder from the original transformer architecture. During pre-training, it uses masked language model (MLM) as the training objective. This allows better comprehension by fusing left and right context. As a result, it is highly suitable for tasks like NER, QA, POS tagging and classification. This model was pre-trained on 104 languages including Marathi.

The size of the BERT model is determined by its parameters, viz: $A$ (the number of Attention Layers), $L$ (the number of Encoder Layers), and $H$ (the number of Hidden Layers). $\textbf{BERT}_{BASE}$ (L=12, H=768, A=12) has 110M parameters and $\textbf{BERT}_{LARGE}$ (L=24, H=1024, A=16) has 340M parameters. (Turc et al., 2019) showed that when we beginning with a pre-trained compact model, transferring task knowledge from large ne-tuned models through distillation (Hinton et al., 2015) results in a simple, yet effective, and general algorithm - *Pre-trained Distillation*, bringing further improvements. Authors released 24 pre-trained miniature BERT models publicly. This spurred further research in the area, as more researchers were able to experiment with smaller models. (Sanh et al., 2020) introduced a distilled version of BERT that retains 97% of its language understanding while reducing the model size by 40% and is 60% faster.

RoBERTa (Liu et al., 2019) studied various strategies to improve BERT performance, such as static vs. dynamic masking, larger batch sizes for training, BPE in place of Wordpiece for text encoding. It also used a novel CC-NEWS dataset and released the implementation at https://github.com/facebookresearch/fairseq.

## 6.2 BART

Bidirectional and Auto-Regressive Transformer (BART) (Lewis et al., 2020) is a denoising autoencoder for pretraining sequence-to-sequence models. BART training consists of corrupting the text with an arbitrary noising function. The training objective of the model is to reconstruct the original text. It uses the standard Transformer architecture which can be considered as generalization of BERT (due to the bidirectional encoder), GPT (with the left-to-right decoder), and other pre-training approaches. In general, BART based models perform best when fine-tuned for text generation and comprehension tasks.

# 7 Indic NLP models

This section covers some of the widely used models for Indic NLP applications.

## 7.1 mT5 - multilingual T5

Text-To-Text Transfer Transformer (T5) (Raffel et al., 2019) harnessed the power of transfer learning by converting all text-based NLP problems into a text-to-text format. mT5 (Xue et al., 2021) extended this to build a single model covering 101 languages including major Indic languages. T5 uses Colossal

Clean Crawled Corpus (C4)[8] and claims to achieve SOTA results on many benchmarks for QA, summarization, sentiment analysis, and other NLP tasks. T5 offers various options like T5-Small (60M), T5-Base (220M), T5-Large (770M), T5-3B (3 billion) and T5-11B (11 billion). mT5-XXL (13 billion) variant achieves SOTA performance for QA and comparable to SOTA for NER tasks. However, it has very large deployment cost.

### 7.2 MahaNLP Monolingual Models & Library

L3Cube-MahaCorpus (Joshi, 2022) added various monolingual BERT based models for Marathi. These models are trained on Marathi monolingual dataset based on different internet sources. They increased the size of existing monolingual Marathi corpus by adding 24.8M sentences. Their models - MahaBERT, MahaAlBERT, and MahaRoBerta are trained on Marathi corpus having 752M tokens. They have also released MahaFT, Marathi FastText (FT) (Bojanowski et al., 2017) embeddings trained on the full corpus. Authors claim that these models perform well on sentiment analysis, NER and text classification tasks. Authors also released MahaGPT, a GPT (Radford et al., 2019) model trained on the Marathi corpus. Authors claim SOTA performance for headline classification, NER and sentiment analysis tasks and suggest that monolingual models perform better than multilingual models.

L3Cube team released mahaNLP (Magdum et al., 2023) open source library for Marathi NLP[9] applications. It improves upon various existing NLP libraries like iNLTK (Arora, 2020), spaCy (Honnibal et al., 2020) and Stanza (Qi et al., 2020). It is based on L3Cube-MahaCorpus. The toolkit offers a comprehensive array of NLP tasks, include fundamental preprocessing tasks and advanced NLP tasks like sentiment analysis, NER, hate speech detection, and sentence completion, which are not found in other NLP libraries.

### 7.3 Facebook AI XLM-R Model

Facebook AI released XLM-R (Conneau et al., 2020), a transformer based MLM model for 100 languages. It significantly outperformed the original mBERT model on various cross-lingual (XLM) benchmarks for tasks like NER, MLQA etc. XLM-R performs particularly well for low-resource languages compared to earlier models. Authors released the code and data publicly[10]. Their best model XLM-RoBERTa (279M parameters), is better than mBERT for cross-lingual classification for low-resource languages. Training the same model for more languages leads to better cross-lingual performance (for low-resource languages) only upto a point, after which the overall performance for monolingual and cross-lingual benchmarks degrades. Authors claim to have avoided the *curse of multilinguality* by increasing the model capacity. The model uses a vocabulary of size 256k. They used cleaned CommonCrawls (Wenzek et al., 2019) for training and SentencePiece for multilingual tokenization directly on raw text and did not observe any loss in performance compared to models trained with language specific preprocessing and BPE.

### 7.4 IndicNLP suite and Multilingual Models

IndicNLP (Kakwani et al., 2020) is a comprehensive set of NLP resources, associated models and dataset for 12 Indian languages[11] including Marathi. It includes a large monolingual corpora, pre-trained word embeddings (FT and MUSE), pre-trained language models, multiple NLU evaluation datasets, and IndicGLUE[12] benchmark. The monolingual corpora *IndicCorp* has 2.31M news articles, 34M sentences and 551M tokens for Marathi from various classes. FastText (FT)[13] is used for embeddings as it is known to perform well for Morphologically rich languages like Marathi. It includes *IndicBERT* (v1), a pre-trained model based on AlBERT (Lan et al., 2019). Authors trained a single model for all supported languages for utilizing relatedness amongst them. This also helps improve performance for low resource languages. Sentencepiece (Kudo and Richardson, 2018) was used for tokenization. It was trained on IndicCorp corpora. MLM objec-tive was used during pre-training with exponentially smoothed weighting of data across languages. Authors released *IndicGLUE* benchmark which is useful for Indic NLU tasks. New datasets for tasks like Wikipedia Section-Title Prediction, Article Genre Classification, Headline Prediction, Multiple Choice QA, Winograd NLI[14] and COPA are also included. *IndiaCorp* released as part of this work was a 9-times larger in size compared to OSCAR (Suárez et al., 2019). Fine tuning was done separately for each language on each task in the IndicGLUE benchmark using respective training sets. For challenging tasks like cross-lingual sentence retrieval, this model performs better than XLM-R (Conneau et al., 2020) and mBERT (Devlin et al., 2018).

---

[8] https://github.com/google-research/text-to-text-transfer-transformer
[9] https://github.com/l3cube-pune/MarathiNLP
[10] https://github.com/facebookresearch/xlm
[11] https://indicnlp.ai4bharat.org/pages/indicnlp-resources/
[12] https://huggingface.co/spaces/evaluate-metric/indic_glue
[13] https://fasttext.cc/

### 7.5 IndicBART

IndicBART (Dabre et al., 2022) is a multilingual, pre-trained auto-regressive model. It supports 11 Indian languages including Marathi. The model has been evaluated for two NLG tasks, viz: NMT and extreme summarization. However, it is suitable for other NLG tasks as well. It is competitive with mBART50 (Liu et al., 2020) despite being significantly smaller. It also performs well on very low-resource translation for languages not used for training in zero-shot setting i.e. without pre-training or fine-tuning. The base model has 244M parameters while *IndicALBART* has just 97M parameters. This makes it highly suitable for deployment in systems with limited resources. It is trained on *IndicCorp* corpora (Kakwani et al., 2020). Authors rely on orthographic similarity between the languages to ensure shared subword vocabulary. Relatedness amongst the languages helps the model to be more compact. For Marathi, IndicBART achieves better ROUGE-L F1 (Lin, 2004) compared to mBART50 for extreme summarization task.

### 7.6 IndicTrans2

Improving over the work done for IndicNLP suite, IndicTrans2 (Gala et al., 2023) added models for all 22 scheduled Indian languages. They released a Parallel Corpus named BPCC, one of the largest publicly available corpora for Indic languages. It contains 230M bitext pairs, out of which 126M were newly added. It includes 644K manually translated sentence pairs. In addition, they released a n-way parallel corpus for all 22 scheduled languages. The sentences are sourced from 16 diverse domains and include everyday conversations in Indian context. They also created **IN22**, a new n-way parallel benchmark covering all major Indian languages. One of the unique features of this benchmark is human translations of sentences taken from India-specific articles from many various domains. IT2 translation models were also released. These include the main English-centric model which has 1.1B parameters and compact variant that has 211M parameters. The Indic-Indic model supports direct translation between all 22 scheduled Indic languages (a total of 462 translation directions) and has 1.2B parameters, and a corresponding compact one has 230M parameters. They used LaBASE (Feng et al., 2022) for sentence embedding. Distilled models are trained with word-level distillation and perform competitively with best IT2 models. The authors claim that IT2 is the better than best Open Source models. Current SOTA (State-Of-The-Art) model is NLLB 54B MOE (Costa-jussà et al., 2022). However, its too large to be to be deployed in practice due to high latency and inference cost (CPU and memory).

---
[14]https://www.tensorflow.org/datasets/catalog/glue#gluewnli

## 8 Evaluation

In order to compare the quality of results for various NLP tasks, automatic evaluation metrics that are close to the human judgement are important. Various automated metrics have been proposed over the years for assessing translation quality in particular. These fall in two broad categories - string-based and model-based. String-based metrics include BLEU (Papineni et al., 2002). chrF (Popović, 2015) and chrF++ (Popović, 2017). Model-based metrics such as BLEURT (Sellam et al., 2020), BERTScore (Zhang et al., 2019), COMET (Rei et al., 2020) and PRISM (Thompson and Post, 2020). However, model-based metrics are limited to languages represented in the underlying pre-trained model. They are trained on human judgment data from a few languages, and their performance for many low-resource languages has not been evaluated. Following sections de-scribe some of the commonly used metrics for evaluating the performance of NMT systems.

## 8.1 BLEU and sacreBLEU

BLEU (Bilingual Evaluation Understudy) (Papineni et al., 2002) is one the earliest and widely used metric to evaluate the quality of machine translation systems. It is designed to calculate similarity based on n-gram overlaps. BLEU scores are highly dependent on tokenization and fail to capture semantic equivalence, which is especially limiting for Indic languages. Secondly, it is suboptimal for low-resource languages. sacreBLEU (Post, 2018) overcomes the limitation of standardization in terms of tokenization to enable fair comparison[15].

## 8.2 chrF++

Character-level n-gram F-scores, chrF was introduced by (Popović, 2015). ChrF++ (Popović, 2017) also incorporates word unigrams and bigrams. Authors claimed that is better correlated with human judgment. It also works well for morphologically rich languages. It uses sacreBLEU to compute the scores. As a result, it is quite popular for Indic NLP tasks.

## 8.3 BERTScore

BERTScore (Zhang et al., 2019) is a model based metric. It calculates a similarity score for each token in the target sentence with each token in the reference sentence. It does this using contextual embeddings based on BERT (Devlin et al., 2018) or similar models. It correlates better with human judgments and provides stronger model selection performance than existing metrics. It is flexible and can be used for NLG tasks like text summarization, paraphrasing and NMT.

## 8.4 COMET

COMET (Rei et al., 2020) is also a model based metric. It uses pre-trained DL models such as XLM-R (Conneau et al., 2020) and relies on semantic similarity and fluency. It was designed to overcome the limitations of word-based metrics that often don't correlate very well with human judgment. It takes into account source input, translation to be evaluated and a reference translation as input. It supports 3 models - Direct Assessments (DA) (Graham et al., 2013), Human-mediated Translation Edit Rate (HTER) (Snover et al., 2006) and metrics compliant with the Multidimensional Quality Metric framework (Lommel et al., 2013). IndicTrans2 (Gala et al., 2023) reports COMET-DA scores for 13 of the 22 Indic languages, including Marathi. The authors also conducted a reference-based evaluation using the COMET-22 DA model (Rei et al., 2022).

## 8.5 ROUGE and BLEURT

Recall-Oriented Understudy for Gisting Evaluation (ROUGE) (Lin, 2004) is an evaluation metric to measure the quality of NLG tasks, especially summarization. It does so by comparing the generated text with a reference/candidate summary. It suffers from problems similar to that of other string-level metrics and does not take semantics into account. Therefore, it can correlate poorly with human judgment. Extensions like ROUGE-WE (Word Embedding) and ROUGE-BERT overcome some of these limitations. Model-based metrics like BLEURT (Sellam et al., 2020) (based on BERT) that can be combined with ROUGE for better results.

# 9 Conclusion

Despite the initial obstacles in NLP research for Indic languages, many of the advances in LLMs did eventually improve the outcomes in medium resource languages like Marathi. Tools and resources built as part of initiatives like XLM-R, NLLB, IndicTrans, mahaNLP have improved the results for NLP tasks like sentiment analysis, NMT, NER, etc. dramatically over the last 5-6 years. However, the morphological richness make it harder for the models to deal with dialect variation, especially for tasks like Text-to-Speech (TTS). Another challenge is that Marathi speakers often use code-mixed Marathi-English or Marathi-Hindi. Tackling this challenge requires the development of code-mixed datasets for better results. Machine Reading Comprehension (MRC) and Summarization tasks remain challenging due to lack of high-quality, annotated MRC and summarization data for Marathi. Cross-lingual Information Retrieval (IR) and Question Answering (QA) is quite limited due to complex sentence structures and limited data. More research in these areas is required.

---

[15] https://github.com/mjpost/sacrebleu

## Acknowledgments

I would like to express my heartfelt gratitude to my advisor, Dr. S. R. Sathe for his guidance and suggestions. I would also like to thank Dr. Ravindra Keskar for his suggestions and review comments.

## References

Gaurav Arora. 2020. iNLTK: Natural Language Toolkit for Indic Languages. *coRR*, pages 66–71.

Dzmitry Bahdanau, Kyunghyun Cho, and Yoshua Bengio. 2015. Neural Machine Translation by Jointly Learning to Align and Translate. *ICLR*.

Paul Baker, Andrew Hardie, Tony McEnery, B D Jayaram, and B S K. 2004. Corpus linguistics and South Asian languages: Corpus creation and tool development. *Literary and Linguistic Computing*, 19(4):509–524.

Pushpak Bhattacharyya. 2004. Building lexical resources for language processing applications. In *Proceedings of the Language Resources and Evaluation Conference (LREC)*, pages 1035–1038.

Pushpak Bhattacharyya. 2010. Indowordnet. In *Proceedings of the Seventh International Conference on Language Resources and Evaluation (LREC'10)*.

Pushpak Bhattacharyya and V. Panchanathan. 2008. Statistical machine translation for indian languages: Mission hindi to marathi. In *Proceedings of the International Conference on Machine Learning and Applications (ICMLA)*, pages 318–323.

Piotr Bojanowski, Édouard Grave, Armand Joulin, and Tomá Mikolov. 2017. Enriching Word Vectors with Subword Information. *Transactions of the Association for Computational Linguistics*, 5:135–146.

Eric Brill. 1992. A simple rule-based part of speech tagger. In *Proceedings of the workshop on Speech and Natural Language*, pages 112–116. Association for Computational Linguistics.

Peter F. Brown, John Cocke, Stephen A. Della Pietra, Vincent J. Della Pietra, Fredrick Jelinek, John D. Lafferty, Robert L. Mercer, and Paul S. Roossin. 1990. A statistical approach to machine translation. *Computational Linguistics*, 16(2):79–85.

Kyunghyun Cho, Bart van Merriënboer, Dzmitry Bahdanau, and Yoshua Bengio. 2014a. On the properties of neural machine translation: Encoder–decoder approaches. In *Proceedings of SSST-8, Eighth Workshop on Syntax, Semantics and Structure in Statistical Translation*, pages 103–111, Doha, Qatar. Association for Computational Linguistics.

Kyunghyun Cho, Bart Van Merriënboer, Caglar Gulcehre, Dzmitry Bahdanau, Fethi Bougares, Holger Schwenk, and Yoshua Bengio. 2014b. Learning Phrase Representations using RNN Encoder-Decoder for Statistical Machine Translation. *EMNLP 2014 - 2014 Conference on Empirical Methods in Natural Language Processing, Proceedings of the Conference*, pages 1724–1734.

Junyoung Chung, Çaglar Gülçehre, KyungHyun Cho, and Yoshua Bengio. 2014. Empirical evaluation of gated recurrent neural networks on sequence modeling. *CoRR*, abs/1412.3555.

Alexis Conneau, Kartikay Khandelwal, Naman Goyal, Vishrav Chaudhary, Guillaume Wenzek, Francisco Guzmán, Edouard Grave, Myle Ott, Luke Zettlemoyer, and Veselin Stoyanov. 2020. Unsupervised Cross-lingual Representation Learning at Scale. In *Proceedings of the 58th Annual Meeting of the Association for Computational Linguistics*, pages 8440–8451, Online. Association for Computational Linguistics.

Marta R Costa-jussà, James Cross, Onur Çelebi, Maha Elbayad, Kenneth Heafield, Kevin Heffernan, Elahe Kalbassi, Janice Lam, Daniel Licht, Jean Maillard, Anna Sun, Skyler Wang, Guillaume Wenzek, Al Youngblood, Bapi Akula, Loic Barrault, Gabriel Mejia Gonzalez, Prangthip Hansanti, John Hoffman, Semarley Jarrett, Kaushik Ram Sadagopan, Dirk Rowe, Shannon Spruit, Chau Tran, Pierre Andrews, Necip Fazil Ayan, Shruti Bhosale, Sergey Edunov, Angela Fan, Cynthia Gao, Vedanuj Goswami, Francisco Guzmán, Philipp Koehn, Alexandre Mourachko, Christophe Ropers, Safiyyah Saleem, Holger Schwenk, and Jeff Wang. 2022. No Language Left Behind: Scaling Human-Centered Machine Translation.

Raj Dabre, Himani Shrotriya, Anoop Kunchukuttan, Ratish Puduppully, Mitesh M. Khapra, and Pratyush Kumar. 2022. IndicBART: A Pretrained Model for Indic Natural Language Generation. *Proceedings of the Annual Meeting of the Association for Computational Linguistics*, 2:1849–1863.

Jacob Devlin, Ming Wei Chang, Kenton Lee, and Kristina Toutanova. 2018. BERT: Pre-training of Deep Bidirectional Transformers for Language Understanding. *NAACL HLT 2019 - 2019 Conference of the North American Chapter of the Association for Computational Linguistics: Human Language Technologies - Proceedings of the Conference*, 1:4171–4186.

Jay Earley. 1970. An efficient context-free parsing algorithm. *Communications of the ACM*, 13(2):94–102.

M. B. Emeneau. 1956. India as a linguistic area. *Language*, 32(1):3–16.

Fangxiaoyu Feng, Yinfei Yang, Daniel Cer, Naveen Arivazhagan, and Wei Wang. 2022. Language-agnostic BERT sentence embedding. In *Proceedings of the 60th Annual Meeting of the Association for Computational Linguistics (Volume 1: Long Papers)*, pages 878–891, Dublin, Ireland. Association for Computational Linguistics.

Philip Gage. 1994. A new algorithm for data compression. In *C Users Journal*, volume 12, pages 23–38.

Jay Gala, Pranjal A. Chitale, Raghavan AK, Varun Gumma, Sumanth Doddapaneni, Aswanth Kumar, Janki Nawale, Anupama Sujatha, Ratish Puduppully, Vivek Raghavan, Pratyush Kumar, Mitesh M. Khapra, Raj Dabre, and Anoop Kunchukuttan. 2023. Indictrans2: Towards high-quality and accessible machine translation models for all 22 scheduled indian languages. *Preprint*, arXiv:2305.16307.

Naman Goyal, Cynthia Gao, Vishrav Chaudhary, Peng-Jen Chen, Guillaume Wenzek, Da Ju, Sanjana Krishnan, Marc'Aurelio Ranzato, Francisco Guzman, and Angela Fan. 2021. The flores-101 evaluation benchmark for low-resource and multilingual machine translation. *Preprint*, arXiv:2106.03193.

Yvette Graham, Timothy Baldwin, Alistair Moffat, and Justin Zobel. 2013. Continuous measurement scales in human evaluation of machine translation. In *Proceedings of the 7th Linguistic Annotation Workshop and Interoperability with Discourse*, pages 33–41, Sofia, Bulgaria. Association for Computational Linguistics.

Ralph Grishman and Beth Sundheim. 1996. Message understanding conference-6: A brief history. In *Proceedings of the 16th conference on Computational linguistics-Volume 1*, pages 466–471. Association for Computational Linguistics.

Geoffrey Hinton, Oriol Vinyals, and Jeff Dean. 2015. Distilling the Knowledge in a Neural Network. *eprint*.

Sepp Hochreiter and Jürgen Schmidhuber. 1997. Long Short-Term Memory. *Neural computation*, 9(8):1735–1780.

Matthew Honnibal, Ines Montani, Sofie Van Landeghem, and Adriane Boyd. 2020. spaCy: Industrial-strength Natural Language Processing in Python.

Girish Nath Jha. 2010. The tdil program and the indian language corpora initiative (ilci). In *Proceedings of the Seventh International Conference on Language Resources and Evaluation (LREC'10)*.

Raviraj Joshi. 2022. L3Cube-MahaCorpus and MahaBERT: Marathi Monolingual Corpus, Marathi BERT Language Models, and Resources.

Divyanshu Kakwani, Anoop Kunchukuttan, Satish Golla, Gokul N.C., Avik Bhattacharyya, Mitesh M. Khapra, and Pratyush Kumar. 2020. IndicNLPSuite: Monolingual Corpora, Evaluation Benchmarks and Pre-trained Multilingual Language Models for Indian Languages. In *Findings of the Association for Computational Linguistics: EMNLP 2020*, volume 3, pages 4948–4961, Stroudsburg, PA, USA. Association for Computational Linguistics.

Kimmo Koskenniemi. 1983. *Two-Level Morphology: A General Computational Model for Word-Form Recognition and Production*. University of Helsinki.

Taku Kudo and John Richardson. 2018. SentencePiece: A simple and language independent subword tokenizer and detokenizer for Neural Text Processing. *EMNLP 2018 - Conference on Empirical Methods in Natural Language Processing: System Demonstrations, Proceedings*, pages 66–71.

Anoop Kunchukuttan, Divyanshu Kakwani, Satish Golla, Gokul N. C., Avik Bhattacharyya, Mitesh M. Khapra, and Pratyush Kumar. 2020. Ai4bharat-indicnlp corpus: Monolingual corpora and word embeddings for indic languages. *Preprint*, arXiv:2005.00085.

Pawan Lahoti, Namita Mittal, and Girdhari Singh. 2022. A Survey on NLP Resources, Tools, and Techniques for Marathi Language Processing. *ACM Transactions on Asian and Low-Resource Language Information Processing*, 22(2).

Zhenzhong Lan, Mingda Chen, Sebastian Goodman, Kevin Gimpel, Piyush Sharma, and Radu Soricut. 2019. ALBERT: A Lite BERT for Self-supervised Learning of Language Representations. *8th International Conference on Learning Representations, ICLR 2020*.

Mike Lewis, Yinhan Liu, Naman Goyal, Marjan Ghazvininejad, Abdelrahman Mohamed, Omer Levy, Ves Stoyanov, and Luke Zettlemoyer. 2020. BART: Denoising sequence-to-sequence pre-training for natural language generation, translation, and comprehension. *Proceedings of the Annual Meeting of the Association for Computational Linguistics*, pages 7871–7880.

Chin-Yew Lin. 2004. ROUGE: A package for automatic evaluation of summaries. In *Text Summarization Branches Out*, pages 74–81, Barcelona, Spain. Association for Computational Linguistics.

Yinhan Liu, Jiatao Gu, Naman Goyal, Xian Li, Sergey Edunov, Marjan Ghazvininejad, Mike

Lewis, and Luke Zettlemoyer. 2020. Multilingual denoising pre-training for neural machine translation. *Transactions of the Association for Computational Linguistics*, 8:726–742.

Yinhan Liu, Myle Ott, Naman Goyal, Jingfei Du, Mandar Joshi, Danqi Chen, Omer Levy, Mike Lewis, Luke Zettlemoyer, Veselin Stoyanov, and Paul G Allen. 2019. RoBERTa: A Robustly Optimized BERT Pretraining Approach. *eprint.*

Arle Richard Lommel, Aljoscha Burchardt, Hans Uszkoreit, Dfki Berlin, and Hans Uszkoreit@dfki De. 2013. Multidimensional Quality Metrics: A Flexible System for Assessing Translation Quality. *eprint.*

Thang Luong, Ilya Sutskever, Quoc Le, Oriol Vinyals, and Wojciech Zaremba. 2015. Addressing the rare word problem in neural machine translation. In *Proceedings of the 53rd Annual Meeting of the Association for Computational Linguistics and the 7th International Joint Conference on Natural Language Processing (Volume 1: Long Papers)*, pages 11–19, Beijing, China. Association for Computational Linguistics.

Vidula Magdum, Omkar Dhekane, Sharayu Hiwarkhedkar, Saloni Mittal, and Raviraj Joshi. 2023. mahaNLP: A Marathi Natural Language Processing Library. *eprint.*

Tomas Mikolov, Kai Chen, Greg Corrado, and Jeffrey Dean. 2013a. Efficient Estimation of Word Representations in Vector Space. *1st International Conference on Learning Representations, ICLR 2013 - Workshop Track Proceedings.*

Tomas Mikolov, Wen-tau Yih, and Geoffrey Zweig. 2013b. Linguistic regularities in continuous space word representations. In *Proceedings of the 2013 Conference of the North American Chapter of the Association for Computational Linguistics: Human Language Technologies*, pages 746–751, Atlanta, Georgia. Association for Computational Linguistics.

George A. Miller. 1995. Wordnet: A lexical database for english. *Communications of the ACM*, 38(11):39–41.

Kishore Papineni, Salim Roukos, Todd Ward, and Wei-Jing Zhu. 2002. BLEU: a Method for Automatic Evaluation of Machine Translation. *Proceedings of the 40th Annual Meeting of the Association for Computational Linguistics*, pages 311–318.

Jeffrey Pennington, Richard Socher, and Christopher Manning. 2014. GloVe: Global vectors for word representation. In *Proceedings of the 2014 Conference on Empirical Methods in Natural Language Processing (EMNLP)*, pages 1532–1543, Doha, Qatar. Association for Computational Linguistics.

Matthew E. Peters, Mark Neumann, Mohit Iyyer, Matt Gardner, Christopher Clark, Kenton Lee, and Luke Zettlemoyer. 2018. Deep contextualized word representations. In *Proceedings of the 2018 Conference of the North American Chapter of the Association for Computational Linguistics: Human Language Technologies, Volume 1 (Long Papers)*, pages 2227–2237, New Orleans, Louisiana. Association for Computational Linguistics.

Maja Popović. 2015. chrF: character n-gram F-score for automatic MT evaluation. In *Proceedings of the Tenth Workshop on Statistical Machine Translation*, pages 392–395, Lisbon, Portugal. Association for Computational Linguistics.

Maja Popović. 2017. chrF++: words helping character n-grams. In *Proceedings of the Second Conference on Machine Translation*, pages 612–618, Copenhagen, Denmark. Association for Computational Linguistics.

Matt Post. 2018. A call for clarity in reporting BLEU scores. In *Proceedings of the Third Conference on Machine Translation: Research Papers*, pages 186–191, Belgium, Brussels. Association for Computational Linguistics.

Peng Qi, Yuhao Zhang, Yuhui Zhang, Jason Bolton, and Christopher D. Manning. 2020. Stanza: A python natural language processing toolkit for many human languages. *Preprint*, arXiv:2003.07082.

Alec Radford, Jeffrey Wu, Rewon Child, David Luan, Dario Amodei, and Ilya Sutskever. 2019. Language models are unsupervised multitask learners. *OpenAI blog*, 1(8):1–24. https://cdn.openai.com/better-language-models/language_models_are_unsupervised_multitask_learners.pdf.

Colin Raffel, Noam Shazeer, Adam Roberts, Katherine Lee, Sharan Narang, Michael Matena, Yanqi Zhou, Wei Li, and Peter J. Liu. 2019. Exploring the Limits of Transfer Learning with a Unified Text-to-Text Transformer. *Journal of Machine Learning Research*, 21:1–67.

Gowtham Ramesh, Sumanth Doddapaneni, Aravinth Bheemaraj, Mayank Jobanputra, Raghavan AK, Ajitesh Sharma, Sujit Sahoo, Harshita Diddee, Mahalakshmi J, Divyanshu Kakwani, Navneet Kumar, Aswin Pradeep, Srihari Nagaraj, Kumar Deepak, Vivek Raghavan, Anoop Kunchukuttan, Pratyush Kumar, and Mitesh Shantadevi Khapra. 2022. Samanantar: The Largest Publicly Available Parallel Corpora Collection for 11 Indic Languages. *Transactions of the Association for Computational Linguistics*, 10:145–162.

Ricardo Rei, José G. C. de Souza, Duarte Alves, Chrysoula Zerva, Ana C Farinha, Taisiya

Glushkova, Alon Lavie, Luisa Coheur, and André F. T. Martins. 2022. COMET-22: Unbabel-IST 2022 submission for the metrics shared task. In *Proceedings of the Seventh Conference on Machine Translation (WMT)*, pages 578–585, Abu Dhabi, United Arab Emirates (Hybrid). Association for Computational Linguistics.

Ricardo Rei, Craig Stewart, Ana C Farinha, and Alon Lavie. 2020. COMET: A Neural Framework for MT Evaluation. *Proceedings of the 2020 Conference on Empirical Methods in Natural Language Processing (EMNLP)*, pages 2685–2702.

Ehud Reiter and Robert Dale. 1997. *Building Natural Language Generation Systems*. Cambridge University Press.

Gerard Salton and Christopher Buckley. 1988. Term-weighting approaches in automatic text retrieval. *Information processing & management*, 24(5):513–523.

Victor Sanh, Lysandre Debut, Julien Chaumond, and Thomas Wolf. 2020. DistilBERT, a distilled version of BERT: smaller, faster, cheaper and lighter. *eprint*.

Thibault Sellam, Dipanjan Das, and Ankur P. Parikh. 2020. BLEURT: Learning robust metrics for text generation. In *Proceedings of the 58th Annual Meeting of the Association for Computational Linguistics*, pages 7881–7892, Online. Association for Computational Linguistics.

Rico Sennrich, Barry Haddow, and Alexandra Birch. 2016. Neural machine translation of rare words with subword units. *54th Annual Meeting of the Association for Computational Linguistics, ACL 2016 - Long Papers*, 3:1715–1725.

Shashank Siripragada, Jerin Philip, Vinay P. Namboodiri, and C V Jawahar. 2020. A multilingual parallel corpora collection effort for Indian languages. In *Proceedings of the Twelfth Language Resources and Evaluation Conference*, pages 3743–3751, Marseille, France. European Language Resources Association.

Matthew Snover, Bonnie Dorr, Rich Schwartz, Linnea Micciulla, and John Makhoul. 2006. A study of translation edit rate with targeted human annotation. In *Proceedings of the 7th Conference of the Association for Machine Translation in the Americas: Technical Papers*, pages 223–231, Cambridge, Massachusetts, USA. Association for Machine Translation in the Americas.

Pedro Javier Ortiz Suárez, Benoît Sagot, and Laurent Romary. 2019. Asynchronous pipeline for processing huge corpora on medium to low resource infrastructures. In *Proceedings of the Workshop on Challenges in the Management of Large Corpora (CMLC-7) 2019: Leipzig, Germany, 20 July 2019*, pages 9–16.

Sachin R. Suryawanshi, D. M. Kulkarni, and P. B. Joshi. 1994. A morphological analyser for marathi. In *Proceedings of the International Conference on Computer Processing of Oriental Languages*, pages 24–27.

Ilya Sutskever, Oriol Vinyals, and Quoc V Le. 2014. Sequence to sequence learning with neural networks. In *Advances in neural information processing systems*, pages 3104–3112.

Brian Thompson and Matt Post. 2020. Automatic machine translation evaluation in many languages via zero-shot paraphrasing. In *Proceedings of the 2020 Conference on Empirical Methods in Natural Language Processing (EMNLP)*, pages 90–121, Online. Association for Computational Linguistics.

Jörg Tiedemann. 2012. Parallel Data, Tools and Interfaces in OPUS. In *Proceedings of the Eighth International Conference on Language Resources and Evaluation (LREC'12)*, pages 2214–2218, Istanbul, Turkey. European Language Resources Association (ELRA).

Iulia Turc, Ming-Wei Chang, Kenton Lee, Kristina Toutanova, and Google Research. 2019. Well-Read Students Learn Better: On the Importance of Pre-training Compact Models. *eprint*.

Ashish Vaswani, Noam Shazeer, Niki Parmar, Jakob Uszkoreit, Llion Jones, Aidan N Gomez, Lukasz Kaiser, and Illia Polosukhin. 2017. Attention Is All You Need. *CoRR*, abs/1706.03762.

Guillaume Wenzek, Marie-Anne Lachaux, Alexis Conneau, Vishrav Chaudhary, Francisco Guzmán, Armand Joulin, and Edouard Grave. 2019. Ccnet: Extracting high quality monolingual datasets from web crawl data. *CoRR*, abs/1911.00359.

Yonghui Wu, Mike Schuster, Zhifeng Chen, Quoc V. Le, Mohammad Norouzi, Wolfgang Macherey, Maxim Krikun, Yuan Cao, Qin Gao, Klaus Macherey, Jeff Klingner, Apurva Shah, Melvin Johnson, Xiaobing Liu, ukasz Kaiser, Stephan Gouws, Yoshikiyo Kato, Taku Kudo, Hideto Kazawa, Keith Stevens, George Kurian, Nishant Patil, Wei Wang, Cliff Young, Jason Smith, Jason Riesa, Alex Rudnick, Oriol Vinyals, Greg Corrado, Macduff Hughes, and Jeffrey Dean. 2016. Google's Neural Machine Translation System: Bridging the Gap between Human and Machine Translation. *eprint*.

Linting Xue, Noah Constant, Adam Roberts, Mihir Kale, Rami Al-Rfou, Aditya Siddhant, Aditya Barua, and Colin Raffel. 2021. mT5: A Massively Multilingual Pre-trained Text-to-Text Transformer. *NAACL-HLT 2021 - 2021 Conference of the North American Chapter of the Association for Computational Linguistics: Human Language Technologies, Proceedings of the Conference*, pages 483–498.

Tianyi Zhang, Varsha Kishore, Felix Wu, Kilian Q. Weinberger, and Yoav Artzi. 2019. Bertscore: Evaluating text generation with BERT. *CoRR*, abs/1904.09675.